\documentclass[twoside]{article}
\usepackage{aistats2e}
\usepackage{amsmath}
\usepackage{amsfonts}
\usepackage{natbib}
\begin{document}

%

%

\twocolumn[

\aistatstitle{Piecewise Linear Multilayer Perceptrons and Dropout}

\aistatsauthor{Ian J. Goodfellow}

\aistatsaddress{ Universit\'e de Montr\'eal } ]

\begin{abstract}
We propose a new type of hidden layer for a multilayer perceptron, and demonstrate
that it obtains the best reported performance for an MLP on the MNIST dataset.
\end{abstract}

\section{The piecewise linear activation function}

We propose to use a specific kind of piecewise linear function as the activation function for
a multilayer perceptron.

Specifically, suppose that the layer receives as input a vector $x \in \mathbb{R}^D$.
The layer then computes presynaptic output $z = x^T W + b$ where $W \in \mathbb{R}^{D \times N}$
and $b \in \mathbb{R}^N$ are learnable parameters of the layer.

We propose to have each layer produce output via the activation function $h(z)_i = \text{max}_{j \in S_i} z_j$
where $S_i$ is a different non-empty set of indices into $z$ for each $i$.

This function provides several benefits:

\begin{itemize}
\item It is similar to the rectified linear units \citep{Glorot+al-AI-2011} which have already proven useful
for many classification tasks.
\item Unlike rectifier units, every unit is guaranteed to have some of its parameters receive some training signal at each update step. This is because the inputs $z_j$ are only compared to each other, and not to 0., so one is always guaranteed to be the maximal element through which the gradient flows. In the case of rectified linear units,
there is only a single element $z_j$ and it is compared against 0. In the case when $0 > z_j$, $z_j$ receives
no update signal.
\item Max pooling over groups of units allows the features of the network to easily become invariant to some aspects of their input. For example, if a unit $h_i$
pools (takes the max) over $z_1$, $z_2$, and $z_3$, and $z_1$, $z_2$ and $z_3$
respond to the same object in three different positions, then 
$h_i$ is invariant to these changes in the objects position.
A layer consisting only of rectifier units can't take the max over features like this; it can only
take their average.
\item Max pooling can reduce the total number of parameters in the network. If we pool with non-overlapping
receptive fields of size $k$, then $h$ has size $N / k$, and the next layer has its number of weight parameters
reduced by a factor of $k$ relative to if we did not use max pooling. This makes the network cheaper to train and
evaluate but also more statistically efficient.
\item This kind of piecewise linear function can be seen as letting each unit $h_i$ learn its own activation function. Given large enough sets $S_i$, $h_i$ can implement increasing complex convex functions of its input. This includes functions that are already used in other MLPS, such as the rectified linear function and absolute value rectification.

\end{itemize}

\section{Experiments}

We used $S_i = \{ 5 i, 5 i + 1, ... 5 i + 4 \}$ in our experiments.
In other words, the activation function consists of max pooling over non-overlapping groups of
five consecutive pre-synaptic inputs.

We apply this activation function to the multilayer perceptron trained on MNIST by
\citet{Hinton-et-al-arxiv2012}. This MLP uses two hidden layers of 1200 units each.
In our setup, the presynaptic activation $z$ has size 1200 so the pooled output of
each layer has size 240.
The rest of our training setup remains unchanged apart from adjustment to hyperparameters.

\citet{Hinton-et-al-arxiv2012} report 110 errors on the test set. To our knowledge, this is
the best published result on the MNIST dataset for a method that uses neither pretraining nor
knowledge of the input geometry.

It is not clear how \citet{Hinton-et-al-arxiv2012} obtained a single test set number. We train
on the first 50,000 training examples, using the last 10,000 as a validation set. We use the
misclassification rate on the validation set to determine at what point to stop training.
We then record the log likelihood on the first 50,000 examples, and continue training
but using the full 60,000 example training set. When the log likelihood of the validation set
first exceeds the recorded value of the training set log likelihood, we stop training the
model, and evaluate its test set error. Using this approach, our trained model made 94 mistakes
on the test set. We believe this is the best-ever result that does not use pretraining or
knowledge of the input geometry.

\bibliography{strings,strings-shorter,ml,aigaion-shorter}
\bibliographystyle{natbib}

\end{document}